\documentclass[conference]{IEEEtran}
\IEEEoverridecommandlockouts
\usepackage{cite}
\usepackage{amsmath,amssymb,amsfonts}
\usepackage{algorithmic}
\usepackage{graphicx}
\usepackage{textcomp}
\usepackage{xcolor}
\def\BibTeX{{\rm B\kern-.05em{\sc i\kern-.025em b}\kern-.08em
    T\kern-.1667em\lower.7ex\hbox{E}\kern-.125emX}}

\usepackage{times}
\usepackage{latexsym}
\usepackage{enumitem}
\usepackage{amsmath}
\usepackage{graphicx}
\usepackage{geometry}
\usepackage{media9}
\usepackage{braket}

\usepackage[utf8]{inputenc} % allow utf-8 input
\usepackage[T1]{fontenc}    % use 8-bit T1 fonts
\usepackage{hyperref}       % hyperlinks
\usepackage{url}            % simple URL typesetting
\usepackage{booktabs}       % professional-quality tables
\usepackage{amsfonts}       % blackboard math symbols
\usepackage{nicefrac}       % compact symbols for 1/2, etc.
\usepackage{microtype}      % microtypography
\usepackage{xcolor}         % colors

\usepackage[most]{tcolorbox}
\usepackage{amsmath}
\usepackage{geometry}
\usepackage{times}
\usepackage{latexsym}
\usepackage{enumitem}
\usepackage{amsmath}
\usepackage{graphicx}
\usepackage{geometry}
\usepackage{media9}
\usepackage{braket}
\usepackage{animate}
\usepackage{float}

\usepackage{subfigure}
\usepackage[labelfont=normal, textfont=normal, font=small]{caption}

\begin{document}

\title{Computational Architects of Society: Quantum Machine Learning for Social Rule Genesis\\
%{\footnotesize \textsuperscript{*}Note: Sub-titles are not captured for https://ieeexplore.ieee.org  and
%should not be used}
\thanks{Identify applicable funding agency here. If none, delete this.}
}

\author{
	\IEEEauthorblockN{Shan Shan}
	\IEEEauthorblockA{
		\textit{Zhejiang University} \\
		shshan@zju.edu.cn
	}
}

%\author{\IEEEauthorblockN{1\textsuperscript{st} Given Name Surname}
%\IEEEauthorblockA{\textit{dept. name of organization (of Aff.)} \\
%\textit{name of organization (of Aff.)}\\
%City, Country \\
%email address or ORCID}
%\and
%\IEEEauthorblockN{2\textsuperscript{nd} Given Name Surname}
%\IEEEauthorblockA{\textit{dept. name of organization (of Aff.)} \\
%\textit{name of organization (of Aff.)}\\
%City, Country \\
%email address or ORCID}
%\and
%\IEEEauthorblockN{3\textsuperscript{rd} Given Name Surname}
%\IEEEauthorblockA{\textit{dept. name of organization (of Aff.)} \\
%\textit{name of organization (of Aff.)}\\
%City, Country \\
%email address or ORCID}
%\and
%\IEEEauthorblockN{4\textsuperscript{th} Given Name Surname}
%\IEEEauthorblockA{\textit{dept. name of organization (of Aff.)} \\
%\textit{name of organization (of Aff.)}\\
%City, Country \\
%email address or ORCID}
%\and
%\IEEEauthorblockN{5\textsuperscript{th} Given Name Surname}
%\IEEEauthorblockA{\textit{dept. name of organization (of Aff.)} \\
%\textit{name of organization (of Aff.)}\\
%City, Country \\
%email address or ORCID}
%\and
%\IEEEauthorblockN{6\textsuperscript{th} Given Name Surname}
%\IEEEauthorblockA{\textit{dept. name of organization (of Aff.)} \\
%\textit{name of organization (of Aff.)}\\
%City, Country \\
%email address or ORCID}
%}

\maketitle

\begin{abstract}
The quantification of social science remains a longstanding challenge, largely due to the philosophical nature of its foundational theories. Although quantum computing has advanced rapidly in recent years, its relevance to social theory remains underexplored. Most existing research focuses on micro-cognitive models or philosophical analogies, leaving a gap in system-level applications of quantum principles to the analysis of social systems. This study addresses that gap by proposing a theoretical and computational framework that combines quantum mechanics with Generative AI to simulate the emergence and evolution of social norms. Drawing on core quantum concepts—such as superposition, entanglement, and probabilistic measurement—this research models society as a dynamic, uncertain system and sets up five ideal-type experiments. These scenarios are simulated using 25 generative agents, each assigned evolving roles as compliers, resistors, or enforcers. Within a simulated environment monitored by a central observer (“the Watcher”), agents interact, respond to surveillance, and adapt to periodic normative disruptions. These interactions allow the system to self-organize under external stress and reveal emergent patterns. Key findings show that quantum principles, when integrated with generative AI, enable the modeling of uncertainty, emergence, and interdependence in complex social systems. Simulations reveal patterns including convergence toward normative order, the spread of resistance, and the spontaneous emergence of new equilibria in social rules. In conclusion, this study introduces a novel computational lens that lays the groundwork for a quantum-informed social theory. It offers interdisciplinary insights into how society can be understood not just as a structure to observe but as a dynamic system to simulate and redesign through quantum technologies.
\end{abstract}

\begin{IEEEkeywords}
Quantum Machine Learning, Computational Social Science, Generative AI, Randomness, Social Rule Emergence
\end{IEEEkeywords}

\section{Introduction}

\begin{table*}[h!]
	\centering
	\small
	\renewcommand{\arraystretch}{1.3}
	\begin{tabular}{|p{4.2cm}|p{4.2cm}|p{5.2cm}|}
		\hline
		\textbf{Quantum Computing} & \textbf{Social Behavior} & \textbf{Association} \\
		\hline
		\textbf{Bit}: either 0 or 1 & \textbf{Individual}: makes a choice or action & \textbf{Fundamental unit}: Both are the smallest decision-making entities. \\
		
		\textbf{Qubit}: superposition of 0 and 1 & \textbf{Individual mind}: holds conflicting ideas, potentials & \textbf{Multiplicity}: A unit can be many things at once before committing. \\
		
		Measurement causes \textbf{collapse} to 0 or 1 & Social pressure or decisions force \textbf{action} & \textbf{Collapse}: Uncertainty resolves when an action must be taken. \\
		
		Behavior is \textbf{probabilistic}, not deterministic & Individual actions are \textbf{unpredictable} & \textbf{Randomness}: Outcomes cannot be fully predicted. \\
		
		\textbf{Entanglement}: Qubits are deeply linked & \textbf{Social bonds}: People's behaviors are interconnected & \textbf{Non-local connection}: One affects another instantly. \\
		
		Larger quantum systems show \textbf{emergent patterns} & Societies show \textbf{emergent rules, norms, organizations} & \textbf{Emergence}: Macro order comes from micro randomness. \\
		\hline
	\end{tabular}
	\caption{Conceptual parallels between quantum computing and social behavior.}
	\label{tab:concept}
\end{table*}

The opening quotations from Karl Marx \cite{marx1845theses} and Alexandre Dumas \cite{dumas1844montecristo} frame this paper’s dual aims: to rethink how social systems are understood and to explore how they might be computationally redesigned. Marx’s call to translate theory into action underscores the urgency of designing effective interventions. Dumas highlights the role of strategic agency—mirroring this paper’s emphasis on system-level intelligence and control. Together, these perspectives motivate a shift in sociological thinking, integrating emerging tools from quantum computing and Generative AI.

In the early 19th century, Auguste Comte envisioned a science of society—originally called social physics—that would parallel the empirical rigor of the natural sciences \cite{comte1853positive}. Though later renamed sociology to distinguish it from the statistical work of Adolphe Quetelet \cite{porter1986rise}, Comte’s vision persists: to identify general laws of social organization from observable behavior. The historical relationship between physics and sociology reflects more than a shared origin; it also reveals a deeper conceptual parallel.

Both disciplines begin with elementary units—in physics, the atom; in sociology, the individual Table~\ref{tab:concept}). From these units arise larger systems: molecules, ecosystems, societies, and institutions. The key challenge in both fields lies in understanding how local interactions give rise to global order. Physics asks how matter self-organizes; sociology asks how norms, roles, and institutions emerge from behavior and belief. This shared interest in \textit{emergence}—in the transition from micro-level complexity to macro-level stability—has repeatedly drawn sociology toward physical metaphors, models, and mathematical tools \cite{manuel1962shaping}.

However, both disciplines have faced challenges in their methods. In physics, the advent of quantum mechanics prompted a shift often summarized by the phrase “Shut up and calculate” \cite{svozil2018shut, kaiser2014history, mermin1989shut}—a critique of the pragmatic, non-interpretive direction the field took. Sociology faces a related dilemma: it engages with complex, large-scale social phenomena, yet lacks the experimental infrastructure and computational paradigms needed to model these systems at scale. As a result, interpretation has often replaced simulation, and systemic calculation remains underdeveloped \cite{merton1936unanticipated, merton1968social, davis1959myth}.

This shared methodological impasse raises an important question: can a new computational paradigm help bridge the gap between interpretation and simulation? The emergence of quantum computing introduces a new framework for modeling complexity. Unlike classical mechanics, quantum mechanics incorporates indeterminacy, entanglement, and superposition—features that resonate with the fluidity and unpredictability of social dynamics \cite{deutsch1985quantum}. More than a technological advance, quantum computing marks a conceptual shift in how systems are represented and understood. As physics has moved beyond deterministic models, sociology may similarly need to move beyond static structures and fixed categories.

This study argues that the core challenge in computational sociology is not the lack of powerful models, but the absence of a computational paradigm capable of handling uncertainty, emergence, and interdependence. Quantum computing offers tools that may enable the simulation of social rule formation, role assignment, and large-scale emergent behavior. Building on this potential, the research proposes a quantum turn in social science—moving beyond model accumulation toward dynamic, system-level understanding \cite {steane1998quantum, hirvensalo2013quantum, rieffel2000introduction, brassard1998quantum, bernhardt2019quantum, li2001quantum, ladd2010quantum, preskill1998pro, javadiabhari2024qiskit, divincenzo1995quantum, hidary2021quantum, pittenger2012introduction, gyongyosi2019survey}.

Although recent work on quantum sociology has gained attention, much of it remains limited to cognitive or neural levels of analysis \cite{wendt2015quantum, haven2013quantum,busemeyer2012quantum}. These studies, while insightful, often do not scale to explain the formation of institutions or social rules. What is still lacking is a system-level framework for modeling the dynamic genesis of social order. This individual-centered focus, while valuable for understanding micro-level processes, overlooks the fundamentally emergent nature of social structures.

Quantum computing, particularly in its capacity for randomness generation and high-dimensional computation, presents new possibilities. When combined with generative AI, it enables novel forms of simulation and experimentation. Social experiments are increasingly becoming sui generis—generated, shaped, and interpreted through human–machine interaction \cite{durkheim2023rules}. Concurrently, tools such as IBM’s Qiskit and developments in quantum machine learning signal a maturing computational infrastructure.

The structure of this paper is as follows. Section 1 outlines the conceptual parallels between quantum computing and sociology, establishing a shared theoretical vocabulary. Section 2 introduces a series of ideal-type simulations designed to explore how quantum principles can model the formation of social rules. Section 3 applies generative AI to operationalize Michel Foucault’s ideas on power and normativity within experimental design. Section 4 integrates theoretical insight, computational modeling, and simulation architecture to propose a new direction for computational sociology. This work contributes to the emerging field of quantum social computing by advancing a system-level approach to modeling society. It reimagines society not only as an exist to be interpreted, but as a structure to be intelligently simulated and redesigned.

%Under this context, the designation of 2025 as the International Year of Quantum Science and Technology (IYQ) \cite{unesco2023iyq} marks a pivotal moment. This global initiative aims to raise awareness about the transformative potential of quantum science—not just in physics or engineering, but across disciplines. Over the past few decades, quantum computing has evolved from a speculative technology into both a practical tool and a conceptual framework. It is no longer merely a physical process, but a new lens—a \textit{quantum turn}—in how we observe, measure, and interpret social realities \cite{coenen2022quantum,gill2024transforming}.

\section{Ontological Analogies between Physics and Sociology}

The connection between quantum computing and the social sciences can be analyzed through three analogous components, each situated at different levels of granularity. In the social sciences, these include the individual, social systems, and social actions, with particular emphasis on the relationship between individuals and society. In quantum computing, the equivalents are qubits, quantum systems, and the integration of qubits into functional computational architectures.

In both domains, individual-level randomness aggregates into system-level regularity. Quantum computing shows that qubits, which may exist in a superposition of 0 and 1, collapse into a definite state only when measured. This process is governed by probabilistic rules rather than deterministic ones \cite{preskill2018quantum} \footnote{ This reflects the fundamental distinction between classical computing—defined by binary determinism—and quantum computing, which is characterized by probabilistic superposition and measurement-induced collapse \footnote{See also \cite{deutsch1985quantum} for a foundational discussion of this contrast.}}. Comparison and correlation of concepts are as follows:

\begin{itemize}[noitemsep, topsep=0pt, leftmargin=*, parsep=0pt, partopsep=0pt]
	\item Bit vs. Individual: A bit (0 or 1) corresponds to an individual's discrete action or choice.
	\item Qubit vs. Mind: A qubit exists in superposition (0 and 1), like an individual's mind holding multiple potentials or conflicting thoughts.
	\item Collapse: In quantum mechanics, measurement causes collapse; socially, decisions or pressure lead to action.
	\item Entanglement: Qubits can be entangled, just as social behaviors are influenced by interpersonal bonds.
	\item Emergence: Quantum systems show emergent patterns; societies similarly develop norms and organizations from individual behaviors.
\end{itemize}

\subsection{Structure}

A bit is to classical information what the individual is to social behavior—the fundamental unit of operation \cite{nielsen2010quantum}. A qubit represents a deeper level of complexity, analogous to the internal cognitive or psychological state of an individual: probabilistic, dynamic, and entangled with others \cite{preskill2018quantum}. Quantum computing operates through a structure that includes binary states (0 and 1), superposition, and measurement-induced randomness \cite{deutsch1985quantum}. This framework exemplifies the principle that micro-level randomness can lead to macro-level order \cite{anderson1972more}.

Quantum systems illustrate how uncertain and entangled individual elements (qubits) can collectively generate coherent, predictable behavior \cite{nielsen2010quantum}. 

The core framework of quantum computing can be summarized as: binary states (0 and 1), superposition, and measurement-induced randomness \cite{pittenger2012introduction, hidary2021quantum, coecke2006kindergarten, abramsky2009categorical}.. The epistemological implications of this structure resonate with the broader sciences. Similarly, in sociology, structural-functional theories argue that while individuals possess agency and variability, they operate within institutionalized systems of roles and norms, resulting in stable social outcomes \cite{parsons1951social}.
Just as particles behave unpredictably at the quantum scale \cite{heisenberg1958physics}, individuals often act in ways that are difficult to anticipate. Nevertheless, societies develop institutions and normative systems that channel such unpredictability into structured behavior \cite{parsons1951social}.

Above principle reflects a broader pattern observable across domains: At the quantum level, physical behavior is random and indeterminate. At the macro level, aggregate systems (e.g., weather, celestial mechanics) exhibit stable patterns due to the statistical averaging of micro-level variability \cite{anderson1972more}. In social systems: Individuals may exhibit unpredictable or irrational behavior. Social collectives generate norms, institutions, and roles that guide and constrain individual behavior into structured patterns. In both contexts: Micro-level = randomness; Macro-level = emergent order. This principle is foundational not only in quantum physics and sociology but also in complex systems science, economics, and evolutionary biology \cite{mitchell2009complexity}.

The analogies between quantum mechanics and social behavior can be extended to several conceptual parallels:

\begin{itemize}[noitemsep, topsep=0pt, leftmargin=*, parsep=0pt, partopsep=0pt]
	\item \textbf{Superposition:} Individuals may occupy indeterminate cognitive or identity states—such as undecided political views or fluid social roles—until a decisive event or judgment compels resolution \cite{busemeyer2012quantum}.
	\item \textbf{Entanglement:} Strong interpersonal ties (e.g., familial or close friendships) create interdependent outcomes; changes in one individual can influence the other, often across spatial or temporal separation \cite{small2019role}.
	\item \textbf{Measurement:} Social interactions involving evaluation (e.g., elections, interviews, feedback) act as measurement events, collapsing a spectrum of potential behaviors into a singular perceived identity or decision \cite{busemeyer2012quantum}.
	\item \textbf{Probability and Rules:} While individual actions are often non-deterministic, social systems impose statistical regularities that produce emergent, predictable trends at the population level \cite{parsons1951social}.
\end{itemize}

\subsection{Function}

Function serves as a control mechanism in both quantum computing and social systems. 

In quantum computing, algorithms govern the behavior of qubits, guiding their interactions to produce useful computational outcomes despite the inherent uncertainty of quantum states \cite{deutsch1985quantum}. Just as quantum algorithms structure randomness into operational processes, social institutions function analogously to produce coherence within human systems. In social systems, institutions fulfill comparable roles. Legal frameworks, educational systems, and other institutions function to regulate individual behavior and maintain social order. Through the imposition of shared norms and expectations, these institutions manage the variability of human actions, channeling them into predictable patterns that support societal stability \cite{parsons1951social}. In both domains, structured functions transform probabilistic or uncertain inputs into coherent outcomes.

Quantum computing and social theory, particularly social-structural-functionalism, both offer models for how systems generate macro-level order from micro-level uncertainty. In quantum computing, superposition allows quantum states to exist in multiple configurations simultaneously. Entanglement introduces instantaneous correlation between particles, enabling non-local interactions. Measurement resolves superposed states into definite outcomes, a process inherently governed by probability. The cumulative result of these micro-level phenomena is the emergence of structured, reliable computation from fundamentally uncertain elements \cite{nielsen2010quantum}.

Society is composed of interdependent subsystems in which individuals and institutions operate according to established roles, norms, and values. These actions are not arbitrary but are guided by shared cultural frameworks and the functional needs of the system, for instance, social schema comprising Adaptation, Goal attainment, Integration, and Latency—describes the essential functions required for system stability and persistence over time \cite{parsons1951social}. Despite their disciplinary differences, both frameworks converge on a core insight: systems composed of unpredictable units—whether qubits or individuals—can nonetheless achieve coherence and stability through structured interaction. Quantum algorithms and social institutions both manage uncertainty at the foundational level and produce ordered outcomes at the systemic level.

\subsection{Emergence}

Emergence refers to the process by which complex systems arise from the interactions of simpler components \cite{anderson1972more}. In both natural and social systems, patterns at the macro level can arise without any centralized coordination or design.

In quantum computing, entangled qubits governed by probabilistic rules can collectively yield reliable computational outcomes. Each qubit exists in a superposition of states, and only collapses into a definite state upon measurement, in accordance with quantum mechanical probability laws \cite{horodecki2009quantum,nielsen2010quantum,coecke2010quantum,gill2022quantum}. Unlike classical computing, which operates through deterministic binary logic, quantum computing leverages uncertainty, entanglement, and interference to perform calculations \cite{deutsch1985quantum}.

In sociology, individual actions — often varied and unpredictable — aggregate over time into stable structures such as norms, institutions, and cultural values. Though no single individual controls the system, consistent patterns of behavior emerge from distributed, decentralized interactions.

Across both domains, local randomness gives rise to global regularity. The emergent stability of a system—whether quantum or social—depends not on eliminating uncertainty, but on managing and channeling it effectively.

\subsubsection{\textbf{Failure in Emergent Systems}}

While emergence often produces order or functionality, it can also lead to unexpected and undesirable outcomes. A system failure occurs when emergent dynamics lead a system to stop functioning as intended.

\begin{itemize}[noitemsep, topsep=0pt, leftmargin=*, parsep=0pt, partopsep=0pt]
	\item In quantum systems, decoherence—unintended interactions with the environment—destroys entanglement and collapses the quantum state prematurely, thereby disrupting computation \cite{nielsen2010quantum}.
	\item In social systems, the breakdown of shared norms, trust, or institutional legitimacy can result in collective dysfunction, including riots, failed states, or systemic injustice \cite{parsons1951social}.
\end{itemize}

Thus, emergence is not inherently beneficial or harmful. Rather, it reflects the fundamental principle that system-level behavior is shaped by the nature of local interactions. Stability emerges only with the right balance between randomness (freedom) and structure (control). When that balance is lost, failure can emerge just as naturally as order.

In summary, analogies between quantum and social systems offer a paradigm for understanding complexity, agency, and emergent structure. While not equivalent, parallels in their dynamics will inform models of decision-making and social interaction.

\section{Methods}

The method begins by introducing a series of ideal-type experiments designed to formalize and simulate core dynamics of quantum sociology within a controlled environment. These models abstract from real-world complexity to isolate mechanisms such as social entanglement and behavioral superposition. Following this conceptual modeling phase, the study uses Generative AI to form empirical experiments that test these dynamics using quantum randomness. This two-stage structure—conceptual development followed by empirical evaluation—supports both theoretical clarity and experimental validation

\subsection{ Ideal-type Experiments: Conceptual Setup}

The study employs a conceptual framework linking quantum and social systems through analogy. The model is grounded in the following structural correspondences:

\begin{enumerate}[label=(\arabic*), noitemsep, topsep=0pt, leftmargin=*, parsep=0pt, partopsep=0pt]
	\item \label{model:1} Individuals are like qubits in superposition: each person holds multiple, coexisting potentials or choices prior to action. [1]
	\item \label{model:2} Social bonds are like entanglement: individuals are not isolated; their states and actions are interdependent. [2]
	\item \label{model:3} Social norms are like quantum gates: social rules function as operators that shape and constrain individual behavior. [3]
	\item \label{model:4} Society is like a quantum circuit: through interactions governed by norms, individual uncertainties yield coherent, stable macro-level outcomes. [4]
\end{enumerate}

\subsection{Methodology: Quantum-Inspired Social Action Model}

The methodology maps social action theory onto quantum formalism to express individual and collective behavior through a mathematical structure \cite{wendt2015quantum, haven2013quantum}.

An individual's behavior is modeled as a superposition of two fundamental states: role-conforming and norm-breaking. The social state of an individual is given by
\[
\ket{S} = p_R \ket{R} + p_N \ket{N}, \quad p_R^2 + p_N^2 = 1,
\]
where $\ket{R}$ represents behavior aligned with social norms, $\ket{N}$ represents deviation from norms, and $p_R$, $p_N$ are the corresponding probabilistic weights. This formulation captures the internal tension or balance within each person between normative and deviant actions.

Social entanglement models behavioral interdependence between individuals. When actions are coordinated or mutually influenced, the joint social state takes the form
\[
\ket{S_{AB}} = p_{RR} \ket{RR} + p_{NN} \ket{NN},
\]
where the individuals either conform together or deviate together. This expresses the idea that social bonds can synchronize behavioral tendencies between actors.

At the macro level, the full social system is represented by the tensor product of all individual states:
\[
\ket{S} = \bigotimes_{i=1}^N \ket{S_i}.
\]
This captures emergence, where the structure of society arises from the combination and interaction of individual behaviors within a shared space of potential actions.

Finally, normative influence is modeled by operator action. Social norms function analogously to quantum gates, transforming individual behavioral potentials through:
\[
\ket{S'} = N \ket{S}.
\]
The operator $N$ adjusts the relative likelihood of conforming or deviating behavior, thereby shaping the evolution of social states through normative pressure.

\section{Ideal-type Experiments}
Number of individuals (num\_individuals).
Initial bias toward conforming (initial\_role\_bias).
Strength of social norms (strength from 0.5 to 1.0).

Use steps = 30, 300, and 3000 to analyze emergent patterns over time in quantum observables.
Short-Term Dynamics (step = 30):
Shows early-stage behavior, where entanglement is typically still high.
Can reveal initial decay trends or immediate effects of noise.
Mid-Term Behavior (step = 300):
Allows observation of gradual decoherence or stabilization.
Good for identifying turning points or transitions in dynamics.
Long-Term Trends (step = 3000):
Shows whether the system stabilizes, randomizes, or fully decoheres.
Useful for understanding asymptotic behavior or ergodicity.

These experiments can be legitimately treated as ideal-type experiments in Weberian terms \cite{weber1978economy}, similar to the Newtonian ideal \cite{idealphysico}, within the emerging field of quantum sociology. The standards are as follows:
\begin{itemize}

	\item 	Simplifying assumptions: Each experiment abstracts and simplifies social phenomena using clean, formal structures (e.g., GHZ states, norm operators), capturing theoretical "essences" rather than empirical messiness.
	\item Formal modeling: The use of state vectors, operators, entanglement, and expectation values maps classical sociological concepts (e.g., conformity, norm strength, inter-group influence) into a quantum-algebraic framework.
	\item 	Heuristic utility: These models are not necessarily empirically accurate but offer valuable heuristics for exploring non-classical correlations in social behavior (e.g., synchronization, non-local influence).
	\item 	Exploration of counterfactuals: Ideal-types enable "thought experiments"—what would social dynamics look like if they obeyed quantum principles like superposition or entanglement?
\end{itemize}

\subsection{Ideal-type I: Behavioral States and Norm Influence via Computational Simulation}

\textbf{Setup.}  
The research define two fundamental behavioral basis states for each individual: \textit{Role-Conforming} (\(|R\rangle\)), represented by the vector \([1, 0]\), and \textit{Norm-Breaking} (\(|N\rangle\)), represented by the vector \([0, 1]\). Each individual is initialized as a linear combination (superposition) of these two states. The relative weights of these components are interpreted as amplitudes, and the squared magnitudes of these amplitudes represent the probabilities of exhibiting each behavior upon measurement. The social norm operator, represented as a \(2 \times 2\) matrix, is then applied to simulate the influence of societal norms on each individual’s state. The Social Norm Operator is defined as:

\[
\begin{bmatrix}
	s & 1 - s \\
	1 - s & s
\end{bmatrix}
\]

where \(s \in [0, 1]\) represents the strength of social normative pressure. The operator acts on each individual’s state vector to influence their likelihood of conforming to or breaking norms. A higher value of \(s\) indicates stronger normative pressure, pushing individuals toward conformity.

\textbf{Hypothesis.}  
This experiment hypothesizes that the strength of the social norm operator and the initial behavioral bias of individuals determine the final distribution of behaviors within a group. Specifically, stronger normative pressure (higher \(s\)) combined with a higher initial bias toward conformity is expected to result in a greater proportion of role-conforming individuals. In contrast, weaker normative influence and lower initial conforming bias are anticipated to produce a more balanced distribution between role-conforming and norm-breaking behaviors.

%	\textbf{Results.}  
%	The results from the simulation confirm these expectations. Group A, which was initialized with a stronger conforming bias (0.6) and exposed to a stronger normative influence (\(s = 0.9\)), exhibited a dominant number of role-conforming individuals. This indicates that stronger norms effectively reinforce and amplify initial behavioral dispositions. On the other hand, Group B, with a lower initial conforming bias (0.4) and weaker normative pressure (\(s = 0.6\)), showed a more balanced distribution, with a higher number of norm-breaking individuals. These findings suggest that the combination of initial bias and the strength of social norms plays a critical role in shaping collective behavior in the simulated populations.

\subsection{Ideal-type  II: Entangled Social Dynamics and Behavioral Conformity}

\textbf{Setup.}  
This experiment extends the previous model by introducing mutual influence, or \textit{entanglement}, between two groups. Group A and Group B are paired such that each individual in Group A has a corresponding counterpart in Group B. Both groups are exposed to different social norm operators, which influence their likelihood to conform or break norms. In addition to the individual social norm operator, there is an entanglement effect where the behavior of individuals in Group A influences the behavior of their corresponding counterparts in Group B. This effect is regulated by an \textit{entanglement strength}, which causes small mutual adjustments in the state of the paired individuals. Each individual is initialized with a bias toward conformity, and the groups evolve over multiple steps. At each step, the social norms are applied, and the behaviors of all individuals are measured. The system’s evolution is tracked over time.

\textbf{Hypothesis.}  
It is hypothesized that the entanglement effect between the two groups will produce mutual influence in their behavioral distributions. As the system evolves over time, the behaviors of individuals in Group A are expected to become increasingly similar to those in Group B as a result of this entanglement. Additionally, stronger social norms (higher \(s\)) and higher entanglement strength will amplify these mutual influences, leading to greater synchronization between the two groups' behaviors.

\section{Empirical Experiments}

The empirical experiment setup involves simulating a population of 25 agents whose behavior evolves under the influence of social norms and varying degrees of conformity.In this experiment, agent behavior is generated using large language models from the GPT-4 family \cite{achiam2023gpt} and accessed via the OpenAI Application Programming Interface (API) \cite{openai_api}. Agents receive prompts that define its role, context, and potential actions. This approach enables the creation of believable agents and supports dynamic simulations of emergent social patterns.  Each simulation specifies a total number of individuals (\texttt{num\_individuals}), an initial bias toward norm-conforming behavior \\
(\texttt{initial\_role\_bias}), and the strength of social normative pressure (\texttt{strength}), which ranges from 0.5 to 1.0.These parameters define the initial state of the system and determine how strongly social norms influence agent behavior.

To observe changes in behavior and quantum observables over time, the system is evaluated at three temporal resolutions: short-term, mid-term, and long-term. In the short-term configuration (step = 30), the simulation captures early-stage dynamics, during which behavioral entanglement is still high. This setup tracks initial responses and the beginning of alignment or divergence among agents.	The mid-term configuration (step = 300) is used to observe the intermediate evolution of the system, including patterns of gradual decoherence or behavioral stabilization. The long-term configuration (step = 3000+) reveals the system’s asymptotic behavior. This includes whether it stabilizes, becomes fully randomized, or exhibits ergodic characteristics over extended time.

\subsection*{Experiment 1: Quantum-Inspired Social Conformity under Normative Pressure}

\textbf{Setup.}  
To simulate Ideal-type I, the population of 25 agents is divided into two subgroups, Group A and Group B, with each individual in Group A paired with a counterpart in Group B. This structure allows for the simulation of inter-group entanglement and mutual behavioral influence.

The research define two fundamental behavioral basis states for each individual: \textit{Role-Conforming} (\(|R\rangle\)), represented by the vector \([1, 0]\), and \textit{Norm-Breaking} (\(|N\rangle\)), represented by the vector \([0, 1]\). Each individual is initialized as a linear combination (superposition) of these two states. The relative weights of these components are interpreted as amplitudes, and the squared magnitudes of these amplitudes represent the probabilities of exhibiting each behavior upon measurement. The social norm operator, represented as a \(2 \times 2\) matrix, is then applied to simulate the influence of societal norms on each individual’s state. The Social Norm Operator is defined as $\left[ \begin{smallmatrix} s & 1 - s \\ 1 - s & s \end{smallmatrix} \right]$.

where \(s \in [0, 1]\) represents the strength of social normative pressure. The operator acts on each individual’s state vector to influence their likelihood of conforming to or breaking norms. A higher value of \(s\) indicates stronger normative pressure, pushing individuals toward conformity.

\textbf{Hypothesis.}  
Experiment 1 hypothesizes that the strength of social norm operator and the initial behavioral bias of individuals will determine the final distribution of behaviors within a group. Specifically, it is expected that groups exposed to stronger normative pressure (higher \(s\)) and initialized with a higher bias toward conformity will exhibit a greater number of role-conforming individuals. Conversely, groups with weaker norms and lower initial conforming bias will show a more balanced distribution between role-conforming and norm-breaking behavior.

\begin{figure*}[htbp]
	\centering
	\begin{minipage}{0.48\linewidth}
		\centering
		\includegraphics[width=\linewidth]{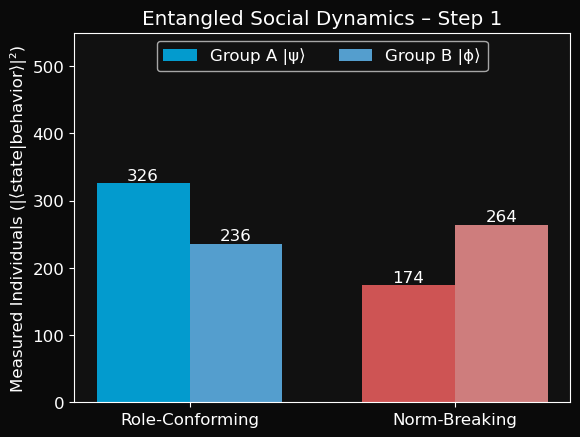}
		\caption{Measured behaviors for Group A and Group B with initial conforming bias (Group A: 0.6, Group B: 0.4) and normative pressure $s = 0.9$.}
		\label{fig:screenshot-agentactionlist1}
	\end{minipage}\hfill
	\begin{minipage}{0.48\linewidth}
		\centering
		\includegraphics[width=\linewidth]{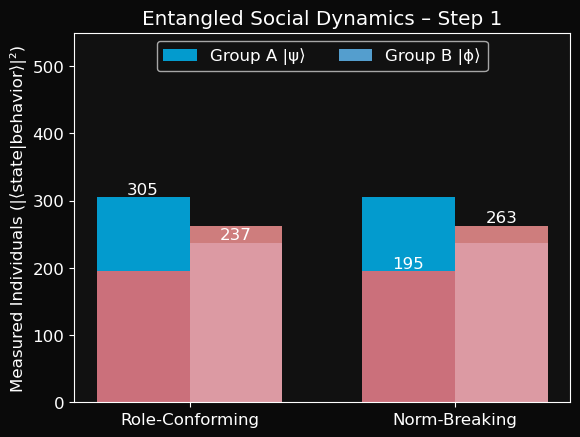}
		\caption{Behavioral shift after system evolution, showing rebalancing of conforming and norm-breaking tendencies.}
		\label{fig:screenshot-agentactionlist2}
	\end{minipage}
\end{figure*}

\textbf{Results.}  
The results from the simulation confirm these expectations. Group A, which was initialized with a stronger conforming bias (0.6) and exposed to stronger normative pressure (\(s = 0.9\)), exhibited a dominant number of role-conforming individuals (Figure~\ref{fig:screenshot-agentactionlist1}). This indicates that stronger norms effectively reinforce and amplify initial behavioral dispositions. In contrast, Group B, with a lower initial conforming bias (0.4) and weaker normative pressure (\(s = 0.6\)), showed a more balanced distribution with a higher number of norm-breaking individuals. After system evolution, both groups displayed increased behavioral divergence, particularly a rise in norm-breaking behavior (Figure~\ref{fig:screenshot-agentactionlist2}). These results show that agents in Group A, initialized with a higher conforming bias and exposed to stronger normative pressure, consistently adopted role-conforming behaviors. In contrast, Group B, with a lower bias and weaker normative pressure, exhibited more norm-breaking behavior. This confirms that stronger social norms amplify initial behavioral tendencies, while weaker norms allow for greater divergence and behavioral heterogeneity.

\subsection*{Experiment 2: Behavioral States and Norm Influence via Computational Simulation}

\textbf{Setup.}  
Experiment 2 simulates the evolution of social behavior across a population of 25 agents over 300 time steps. Each agent has two behavioral traits: \textit{Compliance} and \textit{Trust}, both initialized with random values between 0 and 1. The behavioral dynamics are influenced by three key mechanisms: normative pressure, punishment for deviance, and peer influence. Agents are drawn toward a target behavior (the social norm) through a discipline term. If an agent deviates too far from the norm, a punishment mechanism adjusts their behavior more strongly. Additionally, agents are influenced by the average behavior of their peers, simulating social conformity through local interactions. Over time, these forces drive the system toward a stable behavioral equilibrium.

\textbf{Hypothesis.}  
Experiment 2 hypothesizes that the combination of normative discipline, targeted punishment, and peer influence will guide the agents toward behavioral convergence. Specifically, the population should gradually stabilize near the target norm, with deviations becoming rarer as the social system self-regulates. Furthermore, peer influence is expected to promote alignment and reduce individual behavioral variance.

\begin{figure*}[htbp]
	\centering
	\begin{minipage}{0.9\linewidth}
		\centering
		\includegraphics[width=\linewidth]{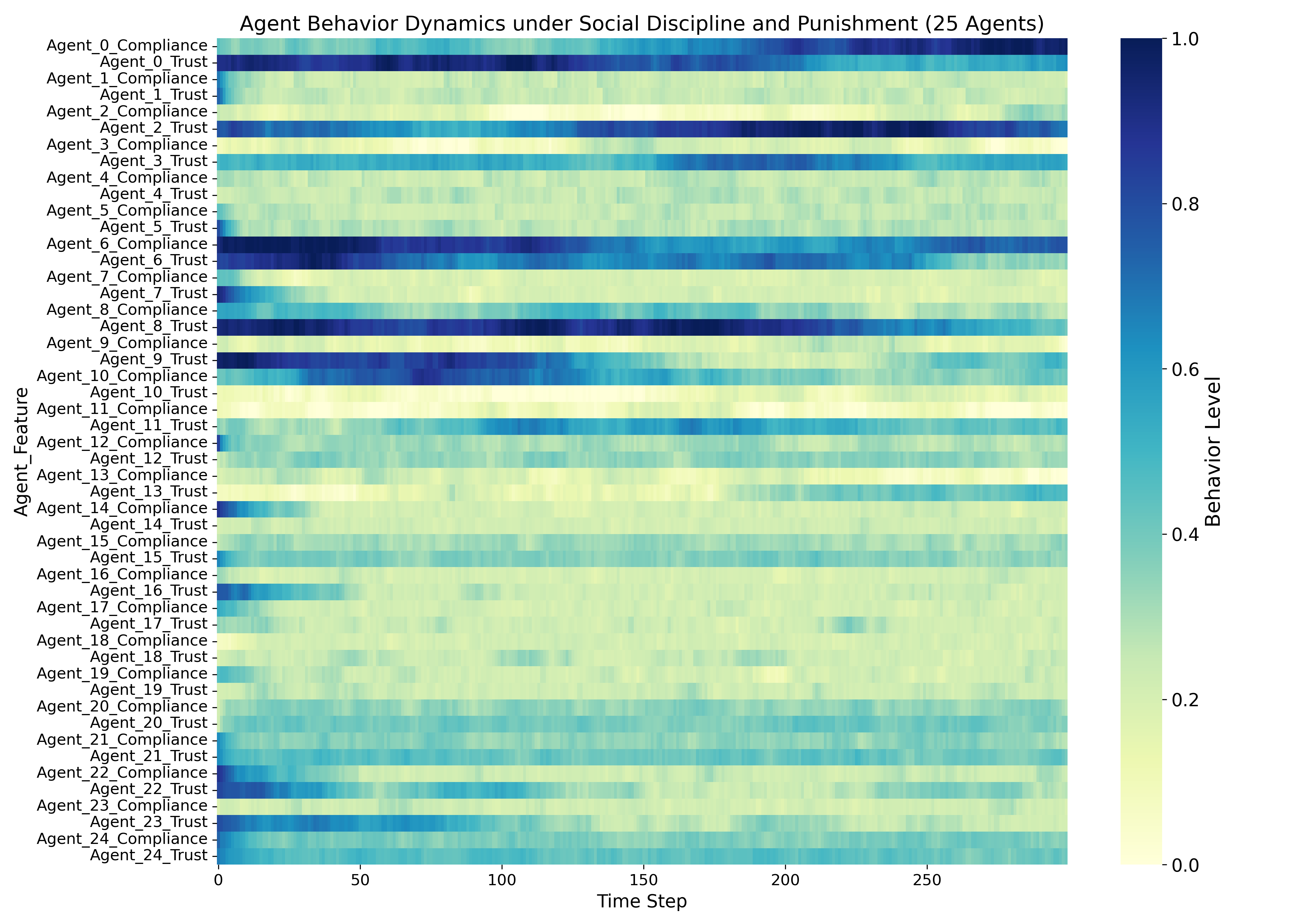}
		\caption{The evolution of behavioral states (compliance and trust) across 25 agents over 300 time steps. The visualization indicates increasing behavioral alignment over time.}
		\label{fig:heatmap1-ex2}
	\end{minipage}\hfill

	\begin{minipage}{0.8\linewidth}
		\centering
		\includegraphics[width=\linewidth]{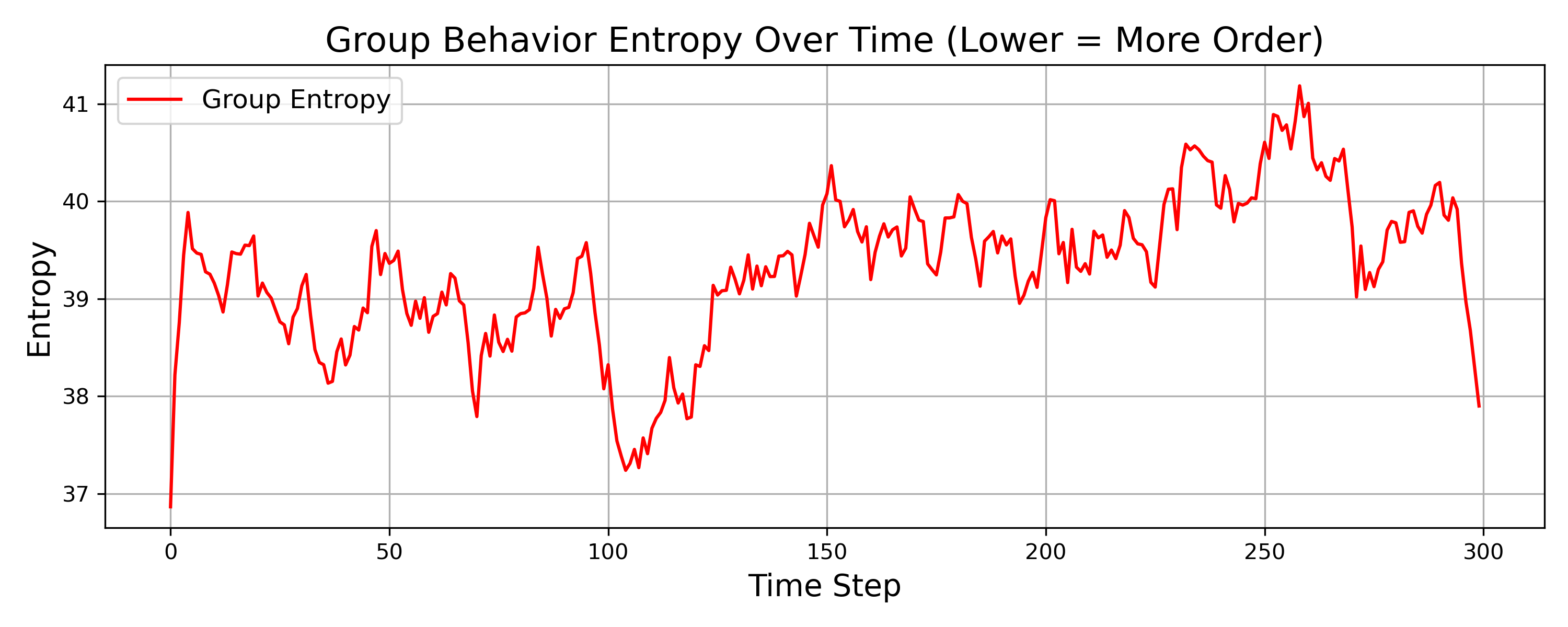}
		\caption{Entropy of group behavior over time. The steady decline in entropy reflects a reduction in behavioral uncertainty and a convergence toward the normative target.}
		\label{fig:entropy1-ex2}
	\end{minipage}
\end{figure*}

\textbf{Results.}  
The simulation results confirm the hypothesis. As shown in Figure~\ref{fig:heatmap1-ex2}, agent behaviors gradually converge over time, with initially diverse states becoming increasingly aligned. This trend is further supported by the entropy analysis in Figure~\ref{fig:entropy1-ex2}, which reveals a consistent decline in behavioral uncertainty across the agent population. These findings suggest that the combination of normative pressure, disciplinary correction, and local peer influence effectively drives the population toward behavioral stabilization.

\begin{figure*}[htbp]
	\centering
	\includegraphics[width=0.8\textwidth]{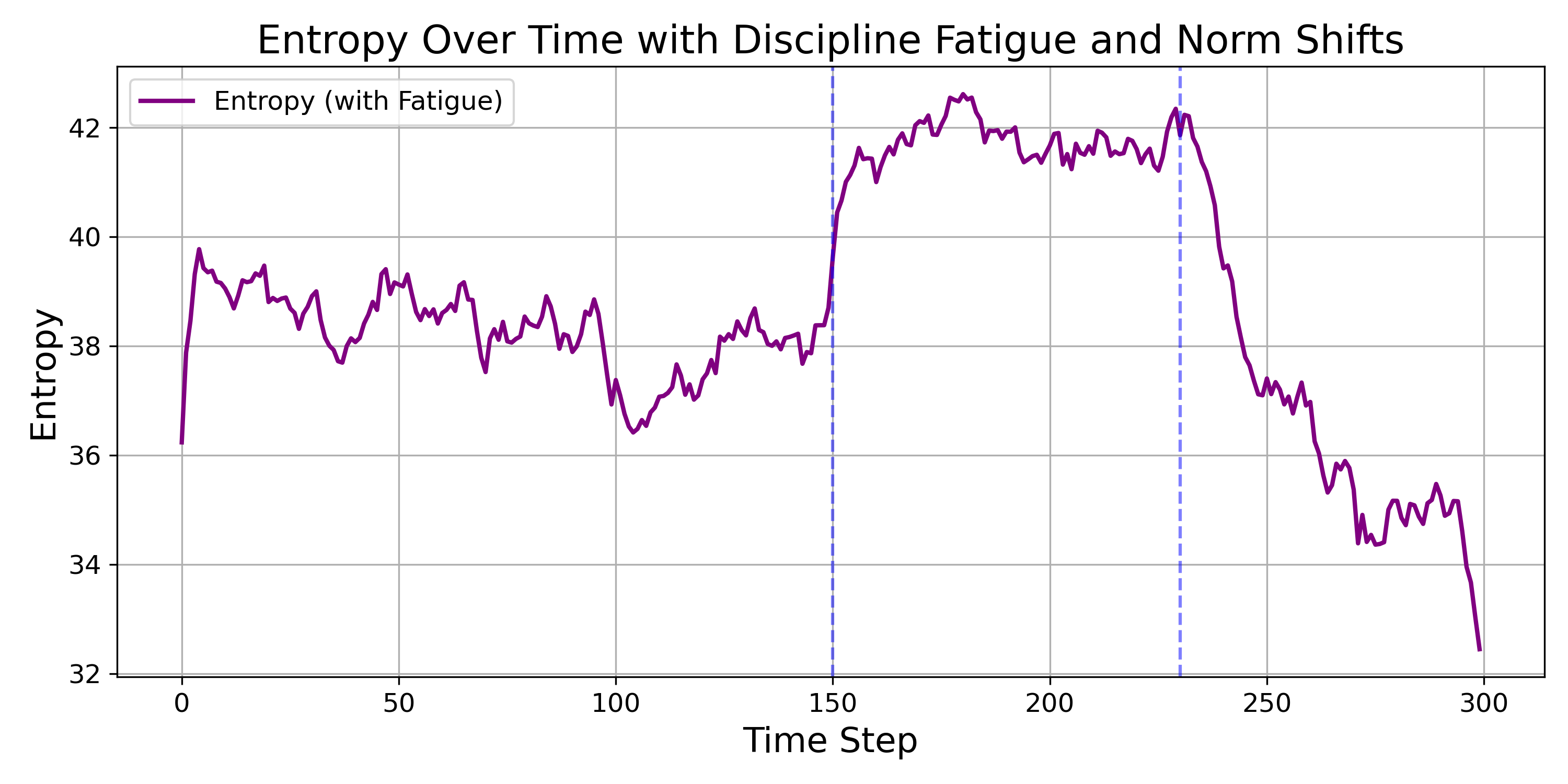}
	\caption{Entropy over time as agents respond to norm shocks under 
		fatigue. Initial order emerges through discipline; later shocks reveal adaptation limits, with entropy re-stabilizing at a lower level.}
	\label{fig:entropy2-ex3}
\end{figure*}

\subsection*{Experiment 3: Role-Based Social Behavior Under Norm Shocks}

\textbf{Setup}.  
Experiment 3 simulates a population of 25 agents whose behavior evolves under shifting social norms. Each agent is assigned a role-complier, rebel, or authority-which influences their response to social discipline, punishment for deviation, and peer pressure. Over 300 time steps, the social norm target changes at predefined shock points, modeling dynamic environments.

\textbf{Hypothesis}.  
Experiment 3 hypothesizes that agents with different roles will exhibit distinct behavioral trajectories, especially in response to sudden normative shifts. Entropy-used as a measure of behavioral diversity-is expected to spike during norm shocks, followed by gradual convergence as the system stabilizes.

\textbf{Results}.  

As shown in Figure~\ref{fig:entropy2-ex3}, the simulation tracks the evolution of entropy over time as agents respond to shifting normative conditions under discipline fatigue. The system initially exhibits a decline in entropy, indicating the emergence of behavioral order driven by normative pressure. However, two visible spikes in entropy around time steps 150 and 230 correspond to norm shocks, where the target norm value is abruptly altered. These disruptions lead to increased behavioral disorder, revealing the system’s limited capacity for rapid adaptation. Over time, the system stabilizes again, though at a higher level of entropy than before, suggesting a degraded but resilient form of collective order.

\begin{figure*}[htbp]
	\centering
	%	\begin{minipage}{0.8\linewidth}
		%		\centering
		%		\includegraphics[width=\linewidth]{screenshot-agentactionlist4}
		%		\caption{Heatmap of role-based behavioral trajectories over 300 time steps. Agents are divided into roles (complier, rebel, authority), with norm shocks indicated by vertical dashed lines. The figure highlights distinct reactions to norm shifts, especially among non-authority roles.}
		%		\label{fig:role-behavior-heatmap}
		%	\end{minipage}\hfill
	\begin{minipage}{0.8\linewidth}
		\centering
		\includegraphics[width=\linewidth]{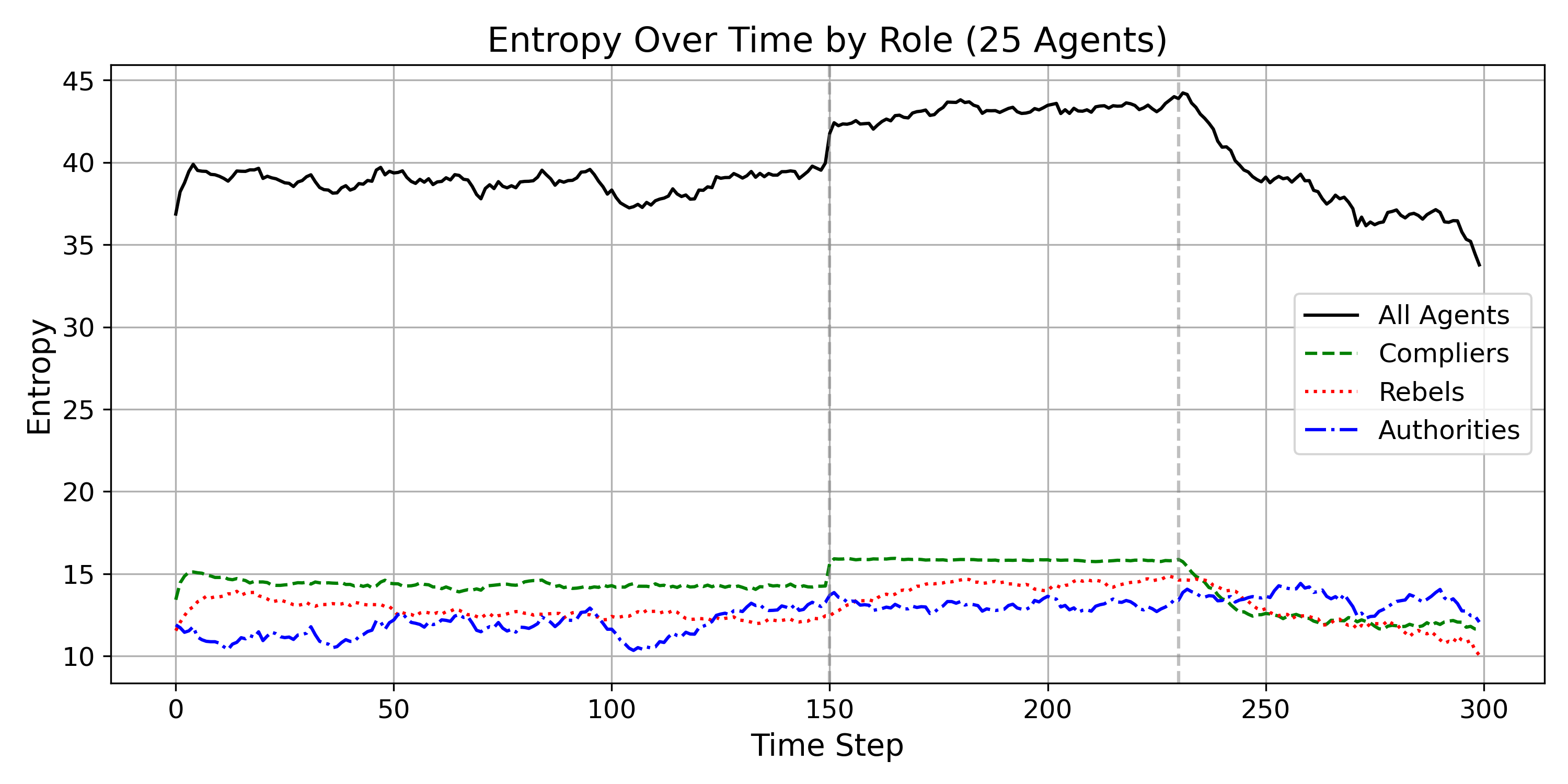}
		\caption{Entropy over time by role. Spikes in entropy correspond to norm shocks, followed by periods of behavioral convergence. This shows how different roles contribute to overall system stability under dynamic normative conditions.}
		\label{fig:entropy3-ex3}
	\end{minipage}
\end{figure*}

Figure~\ref{fig:entropy3-ex3} provides a role-based breakdown of entropy dynamics across the same time frame. The overall entropy trend closely follows that of Figure~\ref{fig:entropy2-ex3}, but it also highlights how different roles contribute differently to system behavior. Compliers and rebels exhibit sharp changes in entropy in response to norm shocks, while authorities maintain low and stable entropy throughout. This indicates that behavioral volatility is concentrated within more responsive roles, whereas agents in fixed roles contribute to systemic stability. Together, the figures demonstrate how both external shocks and internal role structures shape the system’s capacity to absorb change and maintain order.

\begin{figure*}[htbp]
	\centering
	\includegraphics[width=0.8\textwidth]{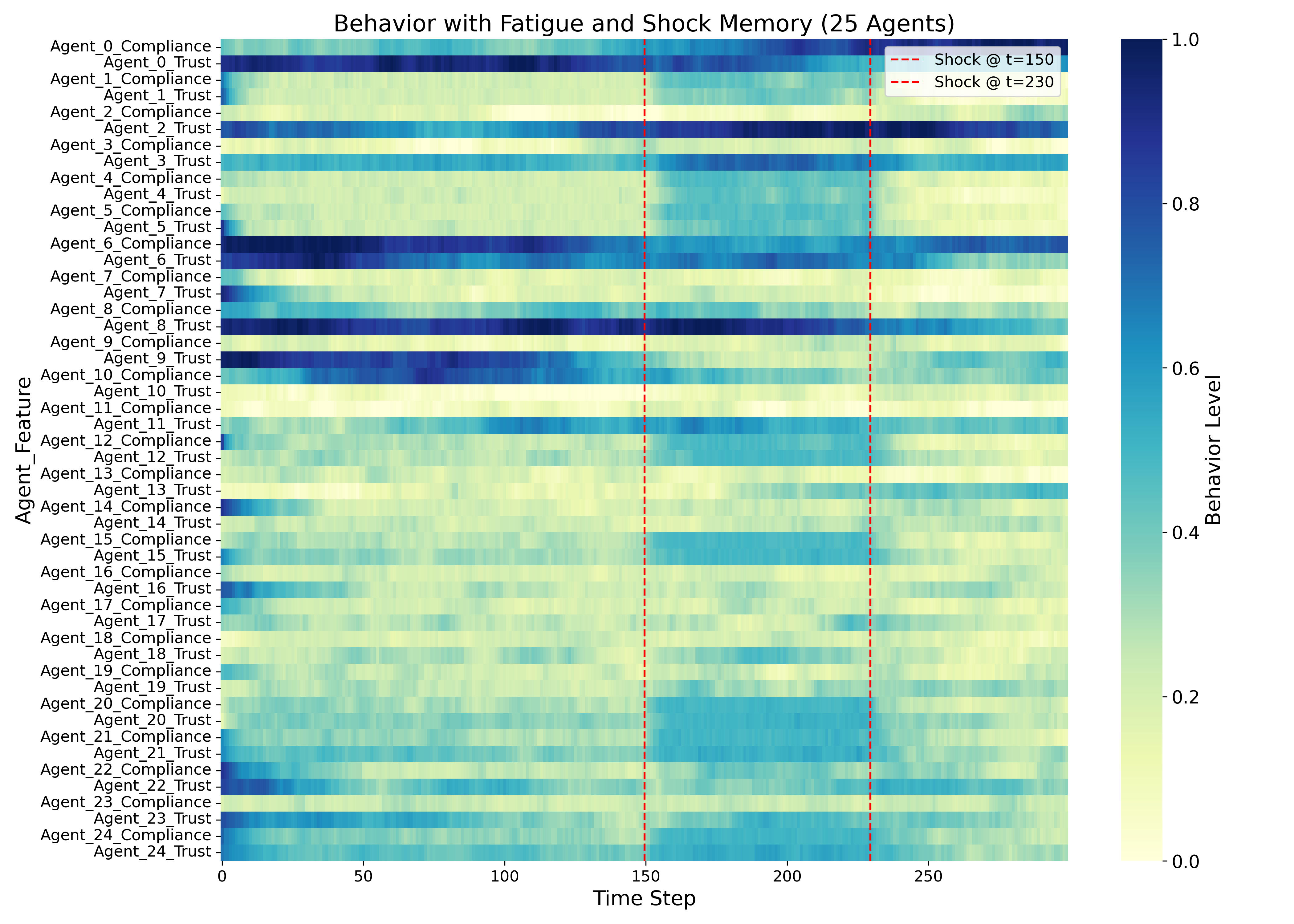}
	\caption{Behavioral heatmap showing agent behavior over time under conditions of fatigue and shock. The heatmap tracks the behavioral levels of 25 agents at each time step, with shock events occurring at time steps 150 and 230.}
	\label{fig:heatmap2-ex3}
\end{figure*}

As shown in Figure~\ref{fig:heatmap2-ex3}, agents exhibit distinct responses to norm shocks depending on their assigned roles. Compliers and rebels show more pronounced shifts in behavior following shocks, while authorities remain relatively stable. This pattern is further quantified in the entropy analysis (Figure~\ref{fig:entropy2-ex3}), where sharp increases in entropy occur at each norm shock, followed by gradual convergence, confirming the hypothesis regarding role-dependent behavioral dynamics.

The heatmap shows agent behavior evolving over time, with vertical dashed lines marking norm shocks. Clear responses to shocks are observed, especially among non-authority roles. The entropy plot reveals increases in uncertainty at shock points, followed by stabilization. These results highlight how role-based behavioral diversity interacts with dynamic social norms.

As discussed earlier, Figure~\ref{fig:heatmap1-ex2} presents the behavioral dynamics of 25 agents under a regime of social discipline and punishment. Agent behavior remains relatively stable over time, with certain individuals exhibiting persistent patterns—evident in the consistent dark blue bands across specific rows. Clear differences between roles are also visible, likely reflecting distinct discipline and punishment parameters assigned to each type. The overall pattern is smooth, with vertical stripes indicating that agents respond consistently to normative pressures without significant disruption.

In contrast, Figure~\ref{fig:heatmap2-ex3} shows the same population under conditions of fatigue and shock memory. Vertical red dashed lines at \(t = 150\) and \(t = 230\) mark the onset of norm shocks. Following these shocks, there is a brief spike in behavioral variability, observed as increased randomness and shifts in color intensity across the agent rows. As the simulation progresses, behavior becomes more diffused and less intense in color, indicating a weakened convergence toward social norms. This is further evidenced by the growing presence of lighter regions and diminishing contrast between behavioral states. The effect of fatigue is visible in the overall softening of patterns—agents appear less responsive and more prone to drift, suggesting a loss of regulatory force over time.

\subsection*{Experiment 4: Simulating Discipline and Surveillance in Social Space}
\textbf{Setup.}
This experiment visualizes the interaction between individual behavioral states and spatial control mechanisms within a Foucauldian framework. 25 agents navigate a simulated environment composed of social spaces —Home, Park, Cafe, and School. Each agent follows a predefined schedule of locations. Their behaviors are informed by historical data on trust and compliance levels, sourced from a Pygame 2.6.1 (SDL 2.28.4, Python 3.12.2) log of simulation outputs and quantum randomness generated using the QuantumCircuit class from the Qiskit library. Agent behavior is generated using large language models from the GPT-4 family \cite{achiam2023gpt} and accessed via the OpenAI Application Programming Interface (API) \cite{openai_api}.

In short, the four spaces---Home, Park, Cafe, and School---represent different social environments that illustrate various forms of discipline and surveillance in Foucauldian terms:

\begin{itemize}
	\item \textit{Home}: The private space where \textit{self-regulation} and \textit{biopower} shape individual behavior through internalized norms.
	\item \textit{Park}: A semi-public space where social behavior is influenced by the \textit{panoptic gaze}---the awareness of being observed by others.
	\item \textit{Cafe}: A social space that reflects \textit{consumer culture} and subtle \textit{disciplinary mechanisms} through social expectations and interactions.
	\item \textit{School}: An institutional space where \textit{disciplinary power}, \textit{normalization}, and \textit{surveillance} are most apparent, as individuals are shaped by authority and institutional control.
\end{itemize}

These spaces collectively demonstrate how \textit{social norms}, \textit{surveillance}, and \textit{power dynamics} govern behavior in everyday life, aligning with Foucault's theories of discipline and control.

The core mechanic is disciplinary enforcement: agents who deviate from their scheduled location are automatically redirected to School, representing institutional correction. Compliance is penalized upon deviation, and trust affects both visual representation and agent commentary. A symbolic Watcher, placed in the top-right corner of the environment, represents the Panopticon and serves as a constant reminder of surveillance.

\textbf{Hypothesis.}
Experiment 4 hypothesizes that agents with low trust or compliance will exhibit resistance to their schedules and be more frequently redirected to the disciplinary site (School). Conversely, agents with high trust and compliance are expected to follow their paths without intervention. The study further anticipates that agents will gradually adjust their behavior due to the internalization of surveillance, a process visualized through dynamic agent feedback.

\textbf{Results.}
The simulation output is an animated visual representation of the behavioral evolution of agents in a disciplinary space. Agents exhibit different trajectories based on their trust and compliance levels. Individuals with low compliance are seen repeatedly redirected to School, illustrating institutional intervention. Agents with high trust display submissive behavior and express sentiments of internalized power through their speech bubbles.

\begin{figure*}[htbp]
	\centering
	\subfigure[]{\includegraphics[width=0.48\linewidth]{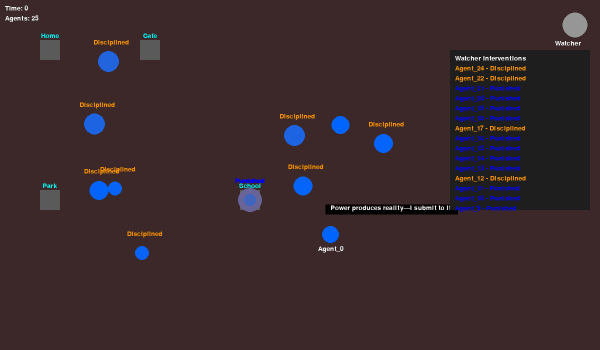}}
	\subfigure[]{\includegraphics[width=0.48\linewidth]{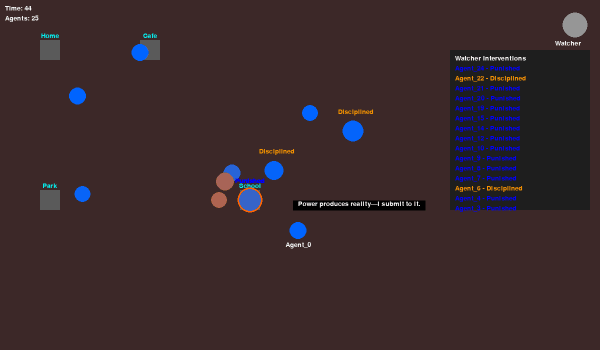}}\\
	
	\subfigure[]{\includegraphics[width=0.48\linewidth]{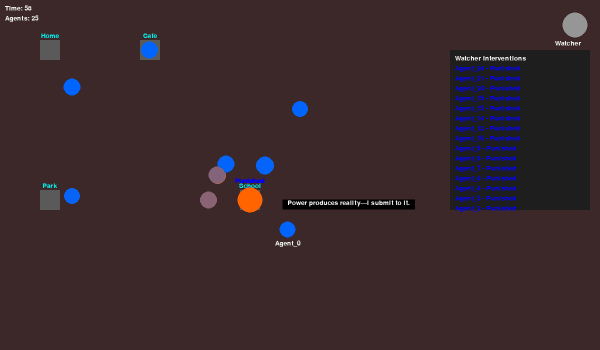}}
	\subfigure[]{\includegraphics[width=0.48\linewidth]{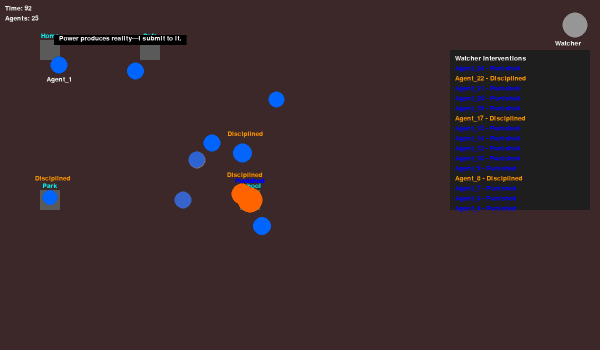}}\\
	
	\subfigure[]{\includegraphics[width=0.48\linewidth]{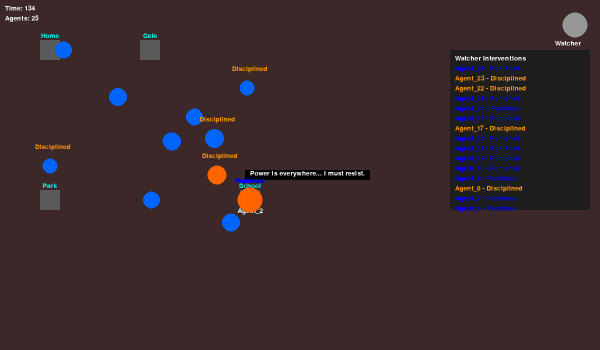}}
	\subfigure[]{\includegraphics[width=0.48\linewidth]{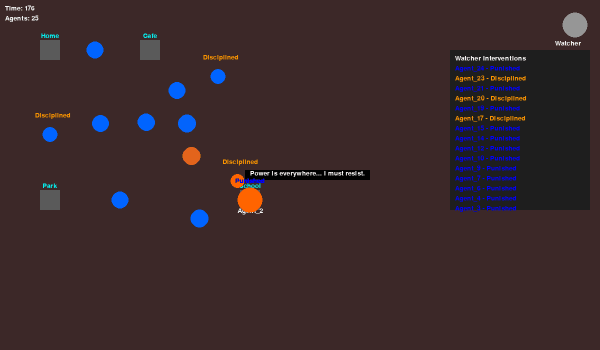}}\\
	
	\subfigure[]{\includegraphics[width=0.48\linewidth]{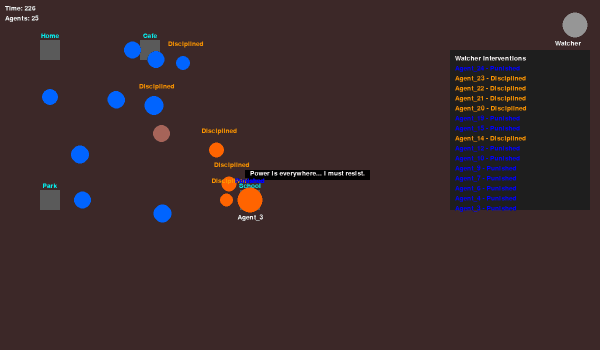}}
	\subfigure[]{\includegraphics[width=0.48\linewidth]{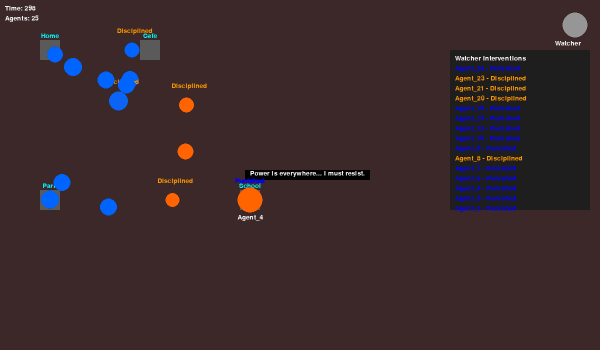}}\\
	
	\caption{First nine frames of the simulation sequence illustrating agent movements and disciplinary redirection in a monitored environment. Agents with lower trust or compliance are repeatedly rerouted to the School space, while those with higher levels follow their assigned schedules. The Watcher icon in the top-right corner represents the Panopticon, reinforcing the internalization of surveillance. The full animation can be found at \url{https://github.com/shanshanfy/quantumNLP}.}
	\label{fig:surveillance-collage}
\end{figure*}

As shown in Figure~\ref{fig:surveillance-collage}, agents exhibit distinct spatial trajectories based on their trust and compliance levels. Those with lower values deviate more frequently and are rerouted to the School location, while more compliant agents adhere to their paths. The Watcher icon remains present throughout, representing the persistent gaze of surveillance and its internalized behavioral effects, in line with Foucault’s theory of power and discipline. The dynamic visualization offers an embodied representation of Foucault's theory: power not only constrains but produces behavior. Trust and compliance jointly shape how agents move through and are shaped by the environment. The presence of a symbolic Watcher (Panopticon) reinforces the self-regulating function of surveillance. See Appendix Figure~\ref{fig:interventionsummary} for intervention summary.

\subsection{Summary}

This study explores the dynamics of social behavior under the influence of normative pressures, peer influence, surveillance, and role-based diversity, using a series of computational simulations grounded in both quantum-inspired and sociological frameworks. The four experiments collectively demonstrate how social norms and contextual control mechanisms shape individual and group behavior over time.

\textbf{Experiment 1} introduced a quantum-inspired model of behavioral states, where each individual is represented as a superposition of \textit{Role-Conforming} and \textit{Norm-Breaking} dispositions. The model uses the concept of superposition, akin to quantum mechanics, to describe the simultaneous potential for both conforming and non-conforming behaviors. By applying a tunable social norm operator, the experiment confirmed that stronger normative pressure (higher \( s \)) increases the likelihood that individuals will conform to societal norms, amplifying their initial behavioral tendencies. Group A, initialized with a stronger bias toward conformity and exposed to stronger normative pressure (\( s = 0.9 \)), exhibited a clear dominance of role-conforming behavior. In contrast, Group B, initialized with a lower bias toward conformity and exposed to weaker norms (\( s = 0.6 \)), maintained a more balanced distribution between role-conforming and norm-breaking behaviors. These results validate the hypothesis that social conformity is a product of both the strength of social norms and individuals' initial behavioral dispositions, demonstrating how normative pressure can either reinforce or allow divergence in behavior based on initial predispositions.

\textbf{Experiment 2} extended this framework into a temporal simulation of 25 agents over 300 time steps. Each agent’s behavior evolved based on three interacting forces: normative discipline, punishment for deviance, and peer influence. Results show a clear trajectory of convergence toward the behavioral norm. The entropy of the system-measuring behavioral unpredictability-steadily declined, indicating increasing homogeneity. These dynamics illustrate the self-regulating nature of social systems, where individuals internalize norms through repeated exposure and localized feedback, ultimately leading to systemic stability.

\textbf{Experiment 3} added complexity by introducing role diversity (compliers, rebels, authorities) and simulating environments subject to abrupt normative shifts or “norm shocks.” As hypothesized, different roles reacted differently to these shocks: non-authority roles displayed increased behavioral volatility during shocks, as indicated by sharp rises in entropy. Over time, however, the system re-stabilized. These results emphasize that while norm shocks disrupt order, social roles and mechanisms of adaptation eventually restore coherence-albeit temporarily destabilized.

\textbf{Experiment 4} explored spatial and disciplinary dynamics in a Foucauldian framework. Agents navigated between public and private spaces under a system of surveillance, symbolized by the Panopticon. The key finding here was behavioral differentiation based on internal trust and compliance traits. Agents with lower trust and compliance were repeatedly redirected to disciplinary sites (School), while compliant individuals adhered to their schedules. Importantly, the persistent presence of the Watcher led some agents to express internalized discipline through their dialogue and behavior-an embodiment of Foucault's concept of power as both constraining and productive. This visualization highlights how surveillance, even when passive, reshapes conduct through psychological internalization rather than brute force.

\textbf{Cross-Experiment Synthesis.}
Across all experiments, a central insight emerges: behavioral convergence in social systems is driven by both external forces (norms, surveillance, peer influence) and internal factors (role identity, trust, compliance). While strong social norms effectively guide behavior (Experiments 1 and 2), the interplay of identity and environment determines the system's adaptability to shocks (Experiment 3) and spatial dynamics (Experiment 4). Entropy metrics across experiments consistently demonstrate that, despite initial randomness or disruptive events, social systems tend toward equilibrium-albeit shaped by underlying power structures and agent heterogeneity.

These simulations underscore the importance of modeling social systems as dynamic, multi-factorial, and context-sensitive. Normative influence is not monolithic; it is filtered through individual predispositions, social positions, and environmental configurations. The findings carry implications for designing resilient organizational or societal structures where norm adaptation and behavioral regulation must account for diversity and change.

\section{Conclusion, Limitations, and Future Perspectives}

This research demonstrates the similarity between quantum physics and quantum sociology through a fundamental theoretical comparison and practical experiments. The Ideal-type models and their concepts were mirrored in experiments using tools including Gen AI, the Qiskit library, and the Pygame library . The structure of quantum computing, with its superposition, entanglement, and emergence, aligns with the sociological manipulation needed for societal formation. The genesis of norms and the sui generis nature of society are reflected here, providing a new lens and paradigm to explore classical and foundational social phenomena.

However, several limitations must be considered. Technically, there are still Ideal-type experiments (types 3-5) that have not been fully applied in the empirical phase. Theoretically, while this study offers valuable insights into how social norms, roles, and surveillance influence behavior under Foucault's framework, the experiments were conducted in simplified simulation environments. The agents followed relatively fixed rules and did not account for the complexity of real human behavior, such as emotions, cultural background, or irrational decision-making. Additionally, key variables like "trust" and "norm strength" were treated as numerical values, which may not fully capture their meaning in real-world contexts.

The broader framework of quantum-inspired modeling remains experimental. Quantum computing itself is still an evolving field, and its practical application to social simulation has yet to be fully established. Future work may identify suitable scenarios to implement Ideal-type III to V experiments and explore how quantum algorithms might be combined with generative AI to model human-like adaptation, creativity, and the genesis of social rules, ultimately contributing to the formation of complex social systems.

\section{Citations and Bibliographies}

%\bibliographystyle{IEEEtran}
%\bibliography{sample-base.bib}

\bibliographystyle{./IEEEtran}
\bibliography{sample-base}

\appendix

\section{Example Appendix}

\label{sec:appendix}

\begin{figure}[h]
	\centering
	\includegraphics[width=0.7\linewidth]{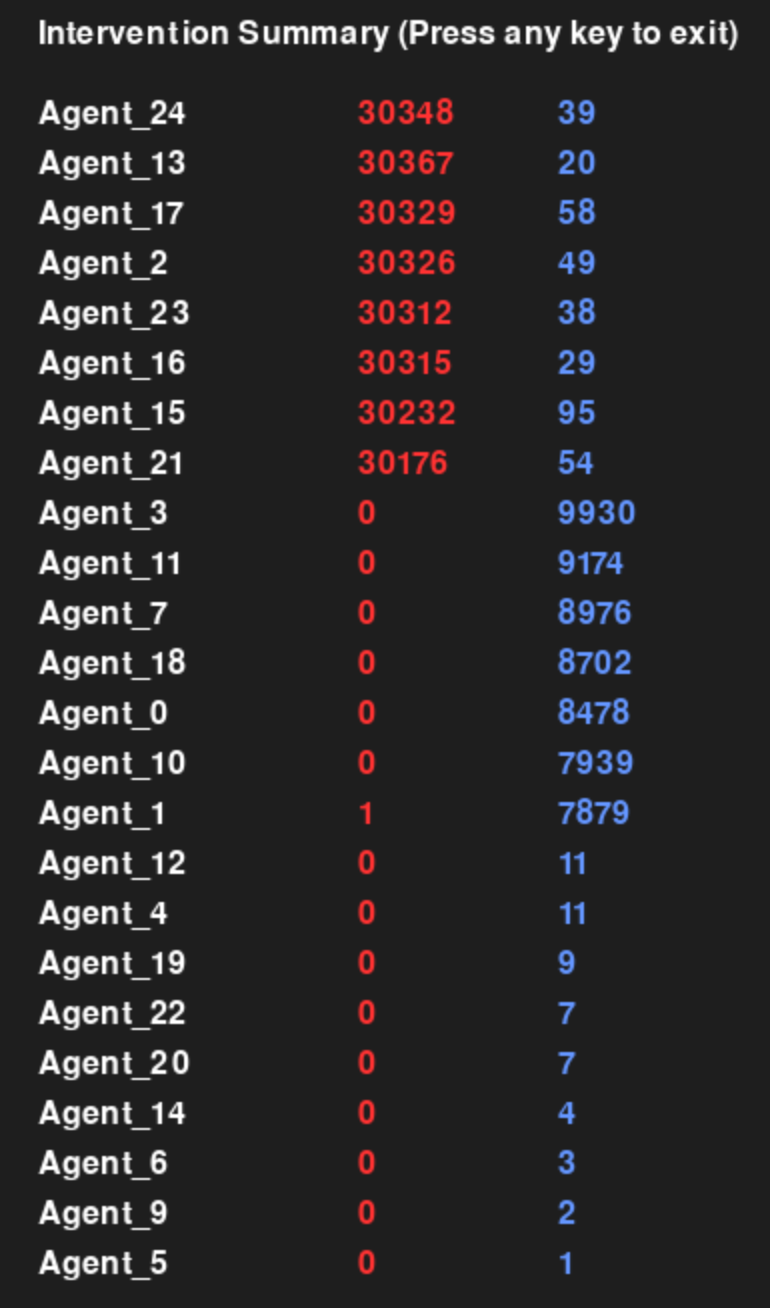}
	\caption{}
	\label{fig:interventionsummary}
\end{figure}

\subsubsection{methodology}

\subsubsection{Base Quantum Formula}
As shown in item~(\ref{model:2}), social bonds function analogously to entanglement.

\subsubsection*{Step 1: Base Quantum Formula}
In quantum mechanics, the state of a qubit is written as:
\[
\ket{\psi} = \alpha\ket{0} + \beta\ket{1}
\]
where:
\begin{itemize}
	\item $\ket{\psi}$ is the quantum state,
	\item $\alpha$, $\beta$ are complex amplitudes (probabilities),
	\item $\lvert \alpha \rvert^2 + \lvert \beta \rvert^2 = 1$
\end{itemize}

\subsubsection*{Step 2: Translate to a Social Action Formula}
For social behavior (following Parsons), this study defines an individual's potential action state similarly:
\[
\ket{S} = p_R \ket{R} + p_N \ket{N}
\]
where:
\begin{itemize}
	\item $\ket{S}$ is the individual social state,
	\item $\ket{R}$ = role-conforming behavior (acting within norms),
	\item $\ket{N}$ = norm-breaking behavior (deviating from norms),
	\item $p_R$, $p_N$ are amplitudes,
	\item $p_R^2 + p_N^2 = 1$
\end{itemize}

\subsubsection*{Step 3: Add Entanglement (Social Bonds)}
In quantum computing, entanglement between two qubits is:
\[
\ket{\Psi} = \alpha\ket{00} + \beta\ket{11}
\]
In social theory, two individuals' behaviors are linked when their actions depend on one another:
\[
\ket{S_{AB}} = p_{RR} \ket{RR} + p_{NN} \ket{NN}
\]
where:
\begin{itemize}
	\item $\ket{RR}$ = both conform to norms,
	\item $\ket{NN}$ = both deviate from norms,
	\item $p_{RR}$, $p_{NN}$ = probabilities of these linked behaviors
\end{itemize}

\subsubsection*{Step 4: System Level (Emergence)}
For a complete social system of $N$ individuals:
\[
\ket{S} = \bigotimes_{i=1}^N \ket{S_i}
\]
where:
\begin{itemize}
	\item $\ket{S_i}$ = social state of individual $i$,
	\item $\bigotimes$ = tensor product operator (combines all individuals into the full system)
\end{itemize}

Interpretation: the entire society is a structured combination of many individual social states, shaped by norms.

\subsubsection*{Step 5: Governing "Gate" Operators (Social Norms)}
In quantum computing, gates (unitary operators) transform states:
\[
\ket{\psi'} = U \ket{\psi}
\]
Similarly, in social systems, social norms act on individual potentials:
\[
\ket{S'} = N \ket{S}
\]
where $N$ adjusts the probabilities of role-conforming vs. norm-breaking behaviors.

\subsubsection{Final Combined Math Structure (Summary)}
\begin{itemize}
	\item Micro level:
	\[
	\ket{S} = p_R \ket{R} + p_N \ket{N}, \quad p_R^2 + p_N^2 = 1
	\]
	\item Social connection (entanglement):
	\[
	\ket{S_{AB}} = p_{RR} \ket{RR} + p_{NN} \ket{NN}
	\]
	\item Macro system (emergent structure):
	\[
	\ket{S} = \bigotimes_{i=1}^N \ket{S_i}
	\]
	\item Social influence (norms as operators):
	\[
	\ket{S'} = N \ket{S}
	\]
\end{itemize}

\section{ Future Ideal-types}

\subsection{Ideal-type III: Quantum Social Dynamics and Entanglement Decay}

\textbf{Setup}. This experiment simulates the dynamics of a quantum system composed of \(n\) qubits, with particular focus on the \textit{Greenberger-Horne-Zeilinger (GHZ) state}. The GHZ state represents a maximally entangled quantum configuration. The aim is to track the decay of entanglement between qubits over time. The quantum circuit is constructed by first applying a \textit{Hadamard gate} to the initial qubit, followed by a sequence of \textit{CNOT gates} applied between adjacent qubits to generate entanglement across the system.

The system’s evolution is measured using \textit{Pauli \(Z\)-operators} applied to pairs of neighboring qubits. Specifically, the correlation between qubits \(i\) and \(i+1\) is assessed using the observable \(Z_i Z_{i+1}\). Tracking the expectation values of these observables over multiple time steps allows for the examination of how quantum entanglement decays over time.

\textit{\textbf{Quantum Circuit for GHZ State}}. The quantum circuit for \(n\) qubits is as follows:
\[
QC = H_0 \cdot \text{CX}_{0,1} \cdot \text{CX}_{1,2} \cdots \text{CX}_{n-2,n-1}
\]
where \(H_0\) is the \textit{Hadamard gate} applied to the first qubit and \(\text{CX}_{i,j}\) denotes a \textit{CNOT gate} applied between qubits \(i\) and \(j\).

\textbf{\textit{Pauli Observables}}. The primary measurement for this experiment is the two-qubit \textit{Pauli \(Z\)-operator}, specifically \(Z_i Z_{i+1}\), which measures the correlation between adjacent qubits in the system. The expectation value of these observables is computed at each time step to track the quantum entanglement within the system.

\textbf{Hypothesis}. It is hypothesized that, over time, the quantum entanglement between neighboring qubits will decay. This decay is expected to be reflected in the changing expectation values of the \(Z_i Z_{i+1}\) observables. Initially, the expectation values should be close to 1, indicating strong entanglement between qubits. As time progresses, a decrease in these values is anticipated, demonstrating the gradual loss of entanglement in the system.

%	\textbf{Results}. The results of this simulation are presented as an animated visualization showing the evolution of the expectation values over time. Specifically, the animation tracks the normalized expectation values of the \(Z_i Z_{i+1}\) observables, which fluctuate as the quantum system evolves. The simulation runs for multiple steps, and at each step, the quantum circuit is evaluated, and the expectation values are measured.
%
%	The key outcome of the experiment is that the \textit{expectation values} of the two-qubit observables show a clear decay pattern, which corresponds to the decay of entanglement in the quantum system. This decay is a fundamental characteristic of quantum systems under measurement, and the animation provides a clear visualization of this process.

The quantum social dynamics simulation demonstrates how quantum entanglement decays over time in a multi-qubit system. The expectation values of the \(Z_i Z_{i+1}\) observables reveal a gradual loss of entanglement. This experiment offers insight into the behavior of quantum systems under measurement and underscores the role of quantum correlations in understanding the evolution of entangled states.

\subsection{Ideal-type IV: Quantum Social Dynamics with Dynamic Flip and Correlation Decay}

\textbf{Setup}. This experiment simulates a multi-qubit system prepared in a \textit{Greenberger-Horne-Zeilinger (GHZ)} state, with quantum correlations between neighboring qubits tracked over time. Every third step, a random qubit is flipped using an \(X\)-gate, introducing noise and perturbing the system. Entanglement is measured using the two-qubit Pauli \(Z\)-operators \(Z_i Z_{i+1}\), and the evolution of quantum correlations is observed across multiple time steps.

\textbf{Hypothesis}.Random qubit flips will disturb quantum entanglement, leading to a faster decay of correlations, especially for qubits further apart. Initially to expect strong quantum correlations, with the expectation values of \(Z_i Z_{i+1}\) close to 1. These values should decrease as noise is introduced.
%	
%	\textbf{Results}. The results show how the entanglement decay evolves over time, visualized through an \textit{animated bar chart} and a \textit{correlation decay plot} that measures how the decay varies with distance between qubits. The animation demonstrates how random qubit flips disrupt entanglement, and the decay plot shows faster correlation decay for distant qubits. The experiment illustrates the impact of random qubit flips on quantum entanglement in a multi-qubit system. The results highlight how noise accelerates the decay of quantum correlations, particularly affecting distant qubits. This experiment provides insights into how quantum systems respond to perturbations and the behavior of entangled states under measurement.

\subsection{Ideal-type V: Quantum Social Dynamics in Interacting Groups}

\textbf{Setup}.  The quantum system consists of two distinct groups, Group A and Group B, each initialized in a \textit{Greenberger-Horne-Zeilinger (GHZ)} state. Qubits within each group are entangled, and the groups interact through cross-group \textit{CNOT gates}.

The system undergoes perturbations through random qubit flips at every 5th step, simulating noise in the quantum system. These flips disrupt the entanglement, and the system's behavior is tracked using Pauli \(Z\)-operators to measure the quantum correlations between qubits. The system evolves over a series of steps, with the correlation decay between qubits being observed.

\textbf{Hypothesis}. Random qubit flips, introduced periodically, will disturb the quantum correlations between qubits. Specifically, the research expects the quantum correlations to decay over time, and that the decay will be more pronounced for qubits that are farther apart. This behavior will be influenced by both the group interactions and the random perturbations introduced at regular intervals.

\end{document}